# A Rigorous Uncertainty-Aware Quantification Framework Is Essential for Reproducible and Replicable Machine Learning Workflows

*Line Pouchard[1,2], Kristofer G. Reyes[1,2], Francis J. Alexander[3], Byung-Jun Yoon[1,4]*

[1]*Brookhaven National Laboratory, Upton, NY 11973, USA*
[2]*University at Buffalo, Buffalo, NY 14260, USA*
[3]*Argonne National Laboratory, IL 60439, USA*
[4]*Texas A&M University, TX 77843, USA*

The ability to successfully replicate predictions by machine learning (ML) or artificial intelligence (AI) models and results in scientific workflows that incorporate such ML/AI predictions is driven by numerous factors, such as the availability of training/testing datasets, the choice of model architectures and parameters, and initial conditions [1,2]. In applications relying on deep learning models, e.g., in image recognition, reproducibility depends upon initializing random seeds that can be silently set by underlying libraries, among other factors [69].  Even when the same data input and initial scripts are re-used, predictions by ML/AI models can exhibit large variability, including outliers that make these results appear unreliable [3,4].  Changing the underlying ML platforms, even new versions of the same, can alter results in a significant way [5,6]. For example, a recent reproducibility study [70] reported that a simple transcription of the same model that was originally implemented in the Java-based Magpie/Weka framework [71] to the Python-based Matminer/scikit-learn framework resulted in a significant unexpected discrepancy in the predictions made by the two platforms. Published results for ML experiments often privilege accuracy obtained with much tuning, and the publications reporting these results may not necessarily provide the ranges of input conditions that produce the reported accuracy, resulting in irreproducible results [7,8].  Varying input ranges for key physical variables in physical experiments and computational studies can be crucial to the applicability of ML algorithms to various classes of experiments. The systems-level view that encourages users to ignore low-level details and focuses instead on modeling the aggregate input-output of a particular process has generated progress in automating experimental and computational scientific workflows.  In this perspective, complicated sub-systems are replaced by black-box ML/AI built from data. However, the probabilistic viewpoint that makes ML powerful at general-purpose modeling can also make its calculations opaque [9,10,11,12,42].  This is particularly important when ML models behave in unpredictable ways or when models are used to predict quantities for which there is no ground truth, as in the cases of models developed for scientific discovery. Instead of a verified result based on trusted calculations, scientists may be faced with varying predictions and no rationale to determine the best course of action.

One of the major challenges scientists will face in the coming years is the integration of ML/AI models and predictions in scientific computational and experimental workflows, whether these predictions replace expensive computational calculations, aid in predicting calculation results, help search through high-dimensional spaces to obtain preliminary candidates for analysis, and numerous new, emerging or yet unforeseen applications. We consider ML/AI predictions for scientific experiments that include numerical simulation campaigns and machine learning tasks, typically orchestrated in computational workflows. Traditionally, scientific workflows rely on building blocks carefully composed with high quality, curated data and first-principles scientific calculations often executed on High Performance Computing (HPC) systems [13]. Examples of promising use of ML/AI in scientific workflows include replacing some computationally expensive modules with cheaper ML-based ones, mitigating challenges that arise from limited and possibly noisy observational data, efficiently sorting through potentially billions of combinations of input candidates in discovery, and aiding just-in-time analysis of high-volume sensor data. To maximize benefits, scientists must be able to replicate the results obtained with these methods, and the ability to measure reproducibility and replicability of computational experiments, scientific workflows, and their outcomes is paramount to establishing trust in the predictions produced by ML/AI models and workflows that incorporate such models.

The attempts to mitigate the problems related to the lack of reproducibility of scientific and computational workflows have generally focused on increasing transparency and settling on a taxonomy of reproducibility. Numerous papers point to the need to increase transparency by providing access to data, code, adequate documentation for methods and execution [14,15,16]; these papers often propose rules and rubrics for measuring the degree to which scientific papers rate on various reproducibility scales [17,18]. To increase transparency, publishers and major conferences have adopted reproducibility requirements for submitted papers [19,20,21]. Containers and software package managers, tools that help build software by capturing dependencies, are often used to satisfy these requirements and provide mitigating solutions [22]. Related to these efforts, the Findable, Accessible, Interoperable, Re-usable (FAIR) principles [23, 24] have provided structured guiding concepts and stimulated tool development for FAIR data and software [25,26,27], including metrics of FAIR compliance [28, 29].

Trustworthy computing has been an active area of research for several decades notably within the National Science Foundation and multiple other federal agencies [30,31]. Trustworthy AI is dealing with complex systems and analytical processing pipelines that raise the bar for trust in computing results [32]. Trustworthy AI requires additional properties to achieve the goal of trustworthy computing: in particular, probabilistic accuracy under uncertainty, fairness, robustness, accountability, explainability, and formal methods are needed [33]. While trust can be subjective, it can be built from such objectified properties, the ability to reproduce results is the property of interest here. The definitions of concepts related to reproducing scientific experiments have been the object of debate among scientists and practitioners, and the preferred taxonomies have not been uniformly adopted across communities [34,35]. The taxonomies proposed by Claerbout [36], Donoho [37] and Peng [38] have informed the definitions proposed in a 2019 report from the National Academies of Sciences, Engineering, and Medicine (NASEM) [39]. NASEM defines *Reproducibility* as "Obtaining consistent computational results using the same input data, computational steps, methods, code, and conditions of analysis". NASEM assigns *Replicability* a broader definition: "Obtaining consistent results across studies aimed at answering the same scientific question, each of which has obtained its own data". In a reversal from its earlier position, the Association for Computing Machinery (ACM) now follows a similar taxonomy to NASEM in its submission policies[1]. In addition, ACM introduces measurements obtained with stated precision and the measuring system. In medical studies and clinical research, the consequences of overfitting for the purpose of raising the statistical significance of a study have been described early [40]. In machine learning for health care "technical, statistical, and conceptual replicability" that are required for full reproducibility of a study involve internal and external validity [41]. While we note that replicability and reproducibility are reversed from the NASEM definitions in [41], the introduction of statistical measures to assess replicability is without doubt a critical initial step in the direction we are proposing for computational workflows.

While the term "reproducibility" is commonly used across diverse science and engineering fields, its meaning is often complex and multi-faceted. An underlying reason is that, while there may be various factors affecting the reproducibility of an experimental outcome or inference result of a scientific workflow, "reproducibility" is frequently used as an umbrella term referring to the net effect of multiple factors affecting the outcome without necessarily differentiating the main sources or factors that contribute to reproducibility, or the lack thereof. It is critical to have the computational means to rigorously quantify the reproducibility of the quantities of interest, as well as assess the respective impact of diverse factors on reproducibility, e.g., stochasticity of the data generation process, potential data corruption (noise or missing values) issues in scientific measurements, model uncertainty, randomness in the model training process. Even when holding fixed these quantities, additional sources of variability exist linked to software, hardware and algorithms [68] - these sources of variability are not addressed here but can be with our approach in a broader perspective.

---

[1] https://www.acm.org/publications/policies/artifact-review-and-badging-current

Key to successfully employing current and future ML/AI methods is quantifying and understanding the uncertainty inherent in their recommendations and predictions. For example, when ML/AI models are employed for decision-making in scientific applications to guide future experiments – where experiments might be costly and time-consuming, experimental resources are limited, or decisions are irreversible – care must be taken in every choice made.

In this article, we    focus on reproducibility (as defined in the NASEM and ACM definitions) as a necessary components of Trustworthy AI. We propose that Uncertainty Quantification (UQ) metrics [45, 46] can be defined within a generalizable, objective-driven, and uncertainty-aware framework to enhance reproducibility for ML/AI. Of specific interest are scientific workflows that involve ML/AI models, whose predictions directly guide or indirectly inform decision-making in the workflow to achieve scientific goals – e.g. discovery, operation, verification.

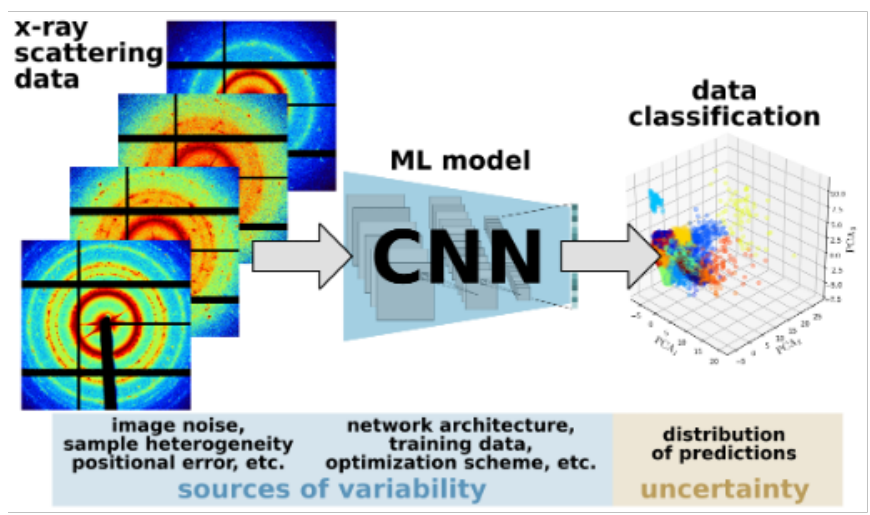

*Figure 1: The complex interactions between sources of variability result in a distribution of predictions that are difficult to interpret (credit: Kevin Yager, Center for Functional Nanomaterials, Brookhaven National Laboratory.).*

For example, consider a relatively simple workflow illustrated in Figure 1, which involves processing x-ray scattering data using a convolutional neural network (CNN) for data classification. The inherent measurement noise and heterogeneity of the training data introduces significant variability in the trained ML/AI model. This gets exacerbated when the sample size of the training data is small with respect to the model complexity, which is common in scientific applications due to the high cost of data acquisition. For example, the design and training of deep network models adds additional variability, where the model architecture, choice of hyper parameters, and the use of popular optimizers based on stochastic gradient descent (SGD) with adaptive learning rates [55,56,57]. The interactions between these sources of variability can be highly complex, partly owing to the non-linear transformations inherent to deep learning models, which contribute to the uncertainty of the ML/AI predictions as well as the final outcomes of the overall workflow. This gives rise to several practical questions. Knowing that there will be inherent uncertainty in the predictions, to what extent can they be trusted? How will the various sources of variabilities and uncertainties affect the reproducibility of the outcomes of a given scientific workflow? How can scientists determine under what conditions they can accept and trust the results obtained from complex workflows that comprise ML/AI models? Clearly, the capability to accurately quantify the reproducibility of the outcomes of a given scientific workflow in the presence of variabilities and uncertainties would be crucial for answering these questions.

In fact, an accurate "uncertainty-aware" metric that can quantitatively assess the reproducibility (or lack thereof) of quantities of interest (QoI) would meaningfully contribute to the trustworthiness of results obtained from scientific workflows involving ML/AI models. Moreover, such a UQ metric will allow us to prioritize the various sources of uncertainties and attribute the (lack of) reproducibility to its primary source, thereby suggesting potential ways to enhance the design and training of the models and the workflow to enhance reproducibility. In addition, it may be used to assess the trade-offs between reproducibility and performance (e.g., prediction accuracy), which will inform researchers to optimize the design and training of the ML models and the overall workflow.

As we elaborate in what follows, UQ in a Bayesian paradigm can provide a general and rigorous quantification framework for reproducibility for complex scientific workflows. It has the potential to fill a critical gap that currently exists in ML/AI for scientific workflows, as it will enable researchers to determine the impact of ML/AI model prediction variability as well as other sources of variabilities on the predictive outcomes of ML/AI-powered workflows. The envisioned framework will contribute to the design of more reproducible and trustworthy workflows for diverse scientific applications, and as a result, ultimately, accelerating scientific discoveries.

## Uncertainty quantification metrics for reproducibility and replicability in ML models

Availability of code and datasets are insufficient metrics to assess reproducibility and replicability for ensemble models and composable workflows as they do not account for the stochasticity of training deep learning models and do not guarantee the reproducibility of the QoI in workflows involving such models.

One promising approach to help design more effective metrics is to consider the distribution of ML/AI prediction results iterated over different training sets and quantify the uncertainty of the predictions (reflected in these distributions). Uncertainty quantification based on a Bayesian paradigm is helpful here – not only because its efficacy in taking these uncertainties into account but also because many factors, including physics-informed quantities/relations and expert knowledge, can contribute to the construction of priors that mathematically represent such uncertainties [58]. Our key hypothesis is that designing metrics that introduce UQ based on a rigorous Bayesian paradigm, will help evaluate the variability of ML predictions when using such algorithms into operations. The process of training and optimizing ML/AI models typically involve various random components, which draw on several sources of variabilities - including the random splitting of available data into training, validation, and test datasets as well as the utilization of stochastic gradient descent (SGD) techniques for model training. This results in randomness of the model predictions, and as a result, when the process is repeated, we obtain an entire probability distribution of model predictions (Fig.1). In this context, *UQ primarily involves understanding and characterizing this distribution as well as quantifying its impact on the predictions of interest*. We can assess existing ML/AI models for the respective impact of diverse factors – stochasticity of the data generation process (e.g., in hydrology [62] and microgrid applications [63]), potential data corruption issues (noise or missing values), model uncertainty, randomness in the model training process, and so forth – on the reproducibility of the results obtained from compositional workflows in operational settings.

Of specific interest is addressing the question of robustness [47,48] in ill- or poorly-posed training of ML/AI models with many parameters, such as deep neural networks. While traditional methods address this through regularization methods [67] through specific terms in a loss function or certain network architectural components such as drop-out regularization, the fundamental issue concerning robustness still remains. Training such models occurs through the optimization of a cost function over a high dimensional space of model parameters, and the optimization methods employed (such as SGD) perform local optimizations without global optimization guarantees. Thus, to mitigate this localness, such optimizations employ heuristics to improve global exploration of the space, such as starting such local optimizations at various, randomly sampled initial points in the parameter space or introducing stochastic perturbations to a search. Such random elements do not guarantee convergence to a global optimum, but stochastic estimates of the optimal parameters with sufficient performance. Studying the stochastic nature of training robustness through a UQ lens does not reduce the uncertainty, or lack of robustness that occurs during training through better optimization techniques or improved regularization. Instead, we quantify such a lack of robustness as a source of uncertainty and understand how such uncertainty impacts a ML/AI model's effectiveness in achieving any experimental objectives for which such a model would be used.

Another aspect of this approach allows understanding reproducibility in the context of the development of models trained on synthetic data: to what degree does model performance transfer when applied to real, experimental data. This workflow – which entails a) the use of physics-based models and simulators to generate synthetic data and the corresponding labels, b) training the ML/AI models on such synthetic data, and c) applying the synthetically-trained model to real world situations – is of special importance to experimental sciences in which obtaining real-world examples and labels is difficult or impossible. Learning how different types of models and their architectures generalize and transfer rules learned from synthetic data to real world experimental data (observations of ground truth) is another benefit of this approach. When a model predicts an observable quantity, this study is relatively straightforward and employing typical metrics to quantify the difference between ML-model predictions and experimental observations may suffice. However, in many cases, models are trained to predict intermediate quantities or quantities that are only partially observable in real-word experiments. For example, we may build a ML/AI model that predicts structure given tomographic data (or otherwise solves an inverse problem). Generating synthetic data to train such a model could involve forward simulations from structure to tomography. However, when applied to physical tomographic data, while a ML/AI model trained on such data may be able to make predictions, it could be hard or impossible to assess to what extent such structural predictions can be trusted. A UQ metric for reproducibility can provide critical insights into the reproducibility of predictions by ML/AI models trained on synthetic data, where variabilities and uncertainties in the design and training process are inevitable. Unless their impact on reproducibility – both of the model predictions and also of the various QoI in the experimental workflow incorporating such predictions – can be rigorously quantified, the role of ML/AI in accelerating scientific discoveries would be significantly hampered.

## Reproducibility and replicability from the perspective of decision-making

Many factors come into play for quantifying uncertainties in ML/AI outcomes for the purpose of designing reproducibility metrics. While ML/AI models may facilitate scientific discoveries in various ways, their ability to effectively assist – and ultimately, automate – decision-making in complex scientific experiments and workflows has an especially strong potential to accelerate scientific advances. For example, ML/AI models can remove the guesswork from experimental design, thereby substantially improving the efficacy of the designed experiments. Furthermore, they may minimize (or eliminate) the need for human intervention in experimental design – an area that still heavily relies on expert intuition – ultimately, enabling autonomous experimental design loops [59, 60]. Without doubt, the reproducibility of the ML/AI predictions that inform or guide "decision-making" in such autonomous experimental loops would be even more crucial. In the context of decision-making, it makes sense to assess reproducibility in terms of how the uncertainties and variabilities in the ML/AI predictions affect the expected efficacy of decision-making.

As we discussed earlier, a UQ metric for reproducibility based on a Bayesian paradigm can provide effective means of quantifying reproducibility (or lack thereof) of various QoI under uncertainty. When the decision-making aspect of ML/AI in scientific workflows is of primary interest, one may define the reproducibility metric based on an "objective-based" UQ framework that quantifies uncertainty based on its impact on decision-making, which will likely suffer due to its presence. For example, the objective-based UQ framework via MOCU (mean objective cost of uncertainty) [43,44] characterizes the model uncertainty by integrating them into a decision-making framework. More specifically, MOCU quantifies the differential cost of decision-making that is expected to increase due to uncertainty, thereby solely focusing on what "actually matters" instead of various other QoI that may be of secondary importance. Due to this focus on optimal and robust decision making under uncertainty, MOCU has been actively applied to optimal experimental design (OED) [49,50,51] and active learning (AL) [52,53,54] in recent years, where the resulting OED and AL strategies have been demonstrated to outperform traditional approaches in goal-driven scientific discoveries. By defining an objective-based UQ metric for reproducibility based on a framework, we can characterize the impact of uncertainties in ML/AI predictions as well as various other sources of variabilities on reproducible decision-making in complex scientific workflows and experiments. Furthermore, such an "objective-driven" reproducibility metric will enable: (i) the identification of sources of uncertainties/variabilities that matter to users; (ii) the measurement of impact on decision-making and its scientific outcomes; and (iii) the design optimization of the model/workflow to enhance reproducibility.

Despite our primary focus on ML/AI-based scientific workflows, the aforementioned approaches can be generally applied to assess reproducibility of scientific workflows and experiments that involve non-AI based modules. These rigorous and general uncertainty-aware frameworks for quantifying reproducibility metrics will be essential in enabling trustworthy ML/AI scientific workflows that produce reproducible outcomes. Such UQ metric for reproducibility can be critical in the development phase of, for instance, climate models that integrate heterogeneous modules developed by a large scientific team of experts and run on different leadership computing facilities (LCF) [61]. In addition, these metrics, when applied with granularity can help identify missing or low-quality data, as well as other potential sources of low reproducibility. Leveraging the decision-making power of frameworks such as MOCU [43, 44] is an effective way of aggregating the impact of uncertainties present in ML/AI models on the reproducibility of end results [5]. Currently, it is practically challenging to accurately identify and attribute the major factors that cause ML/AI model outputs to be highly variable and therefore unreliable. A generic and objective-driven UQ metric for quantifying the reproducibility of ML/AI in the presence of diverse uncertainties in scientific applications can provide the formalism that scientists need to make informed decisions about their choice of models, and parameters given a desired level of reproducibility.

## Potential applications of uncertainty-aware reproducibility metrics

The Bayesian UQ paradigm for uncertainty-aware and objective-based reproducibility quantification as well as the resulting UQ metrics for reproducibility discussed in this article, can play critical roles in enhancing the overall trustworthiness of ML/AI models in scientific workflows. The presented framework provides the means to measure potential trade-offs between accuracy and reproducibility in designing and training ML/AI models, to identify the major sources of variability and uncertainty affecting reproducibility, and to propose potential ways to make the ML/AI predictions and the end results of the workflows incorporating their predictions more reproducible.   While this article did not focus on controls and processes in monitoring the training related to hardware, software, and algorithms, the preliminary inquiries in [6] indicate that our proposed research direction is critical for enhancing reproducibility on future ML/AI platforms.   In a long-term, we envisage various potential applications building on such uncertainty-aware reproducibility metrics, which include: adaptive learning with measuring the reproducibility of predictions when models incorporate pieces of data not seen in training; quantitative evaluation and comparison of ML/AI surrogates in terms of reproducibility; mitigating critical gaps in input data; and automatic model calibration to maximize reproducibility or to optimize the trade-off between accuracy and reproducibility. Finally, we would like to note that there is increasing research efforts to develop UQ methods for ML/AI models, which would play important roles in enabling Bayesian UQ paradigm for uncertainty-aware quantification of reproducibility. While detailed presentation of such methods would be outside the scope of this article, we refer interested readers to relevant papers on the uncertainty quantification of ML/AI models [64, 65, 66, 72, 73, 74], as well as the references therein.

## Acknowledgements

The authors of this manuscript have been supported in part by Brookhaven Science Associates, LLC operator of Brookhaven National Laboratory, a U.S Department of Energy Office of Science laboratory operated under Contract No. DESC0012704.

## Author Contributions

All authors helped to perform the research by conceptualizing ideas and investigating the process and methods to be used; LP, BJY and KR wrote the original manuscript; LP, BJY, and FJA revised the manuscript; FJA supervised the research activity. All authors have reviewed the submission and agreed to the published version of the manuscript.